\documentclass[conference]{IEEEtran}
\IEEEoverridecommandlockouts
\usepackage{cite}
\usepackage{amsmath,amssymb,amsfonts}
\usepackage{algorithmic}
\usepackage{graphicx}
\usepackage{textcomp}
\usepackage{xcolor}
\usepackage{stfloats}

\usepackage[utf8]{inputenc}
\usepackage[T1]{fontenc}
\usepackage{url}
\usepackage{booktabs}
\usepackage{amsfonts}
\usepackage{nicefrac}
\usepackage{microtype}
\usepackage{xcolor}

\usepackage{caption}
\usepackage{subcaption}
\usepackage{multirow}
\usepackage{float}

\usepackage{graphicx}
\usepackage{amsmath}
\usepackage{amssymb}
\usepackage{booktabs}
\usepackage{enumitem}
\usepackage{amsthm}
\usepackage{hyperref}
\usepackage{dsfont}
\usepackage{caption}
\usepackage{comment}
\usepackage{nameref}
\usepackage{enumitem}
\usepackage{soul}
\usepackage{transparent}
\usepackage{xcolor}

\theoremstyle{plain}

\theoremstyle{definition}

\theoremstyle{remark}

\def\BibTeX{{\rm B\kern-.05em{\sc i\kern-.025em b}\kern-.08em
    T\kern-.1667em\lower.7ex\hbox{E}\kern-.125emX}}
\begin{document}

\title{FairAgent: Democratizing Fairness-Aware Machine Learning with LLM-Powered Agents}

\author{\IEEEauthorblockN{Yucong Dai}
\IEEEauthorblockA{\textit{Clemson University} \\
USA \\
yucongd@clemson.edu}
\and
\IEEEauthorblockN{Lu Zhang}
\IEEEauthorblockA{\textit{University of Arkansas} \\
USA \\
lz006@uark.edu}
\and
\IEEEauthorblockN{Feng Luo}
\IEEEauthorblockA{\textit{Clemson University} \\
USA \\
luofeng@clemson.edu}
\and
\IEEEauthorblockN{Mashrur Chowdhury}
\IEEEauthorblockA{\textit{Clemson University} \\
USA \\
mac@clemson.edu}
\and
\IEEEauthorblockN{Yongkai Wu}
\IEEEauthorblockA{\textit{Clemson University} \\
USA \\
yongkaw@clemson.edu}
}

\maketitle

\begin{abstract}
Training fair and unbiased machine learning models is crucial for high-stakes applications, yet it presents significant challenges.
Effective bias mitigation requires deep expertise in fairness definitions, metrics, data preprocessing, and machine learning techniques.
In addition, the complex process of balancing model performance with fairness requirements while properly handling sensitive attributes makes fairness-aware model development inaccessible to many practitioners.
To address these challenges, we introduce FairAgent, an LLM-powered automated system that significantly simplifies fairness-aware model development.
FairAgent eliminates the need for deep technical expertise by automatically analyzing datasets for potential biases, handling data preprocessing and feature engineering, and implementing appropriate bias mitigation strategies based on user requirements.
Our experiments demonstrate that FairAgent achieves significant performance improvements while significantly reducing development time and expertise requirements, making fairness-aware machine learning more accessible to practitioners.
\end{abstract}


\section{Introduction}

Machine learning has revolutionized decision-making across numerous domains, from financial services to healthcare and criminal justice.
However, as these systems increasingly influence crucial decisions affecting people's lives, concerns about algorithmic bias and fairness have come to the forefront of both academic research and public discourse \cite{caton2020fairness, DBLP:conf/aistats/ZafarVGG17, DBLP:conf/www/ZafarVGG17, DBLP:journals/csur/MehrabiMSLG21, DBLP:journals/csur/PessachS23, DBLP:journals/datamine/Zliobaite17, DBLP:journals/widm/QuyRIZN22, wan2022inprocessing}.
Studies \cite{dai2023coupling} have revealed that ML models can perpetuate or even amplify existing societal biases, leading to discriminatory outcomes for certain demographic groups.
This has sparked the development of fairness-aware machine learning \cite{caton2020fairness, DBLP:conf/aistats/ZafarVGG17, DBLP:conf/nips/HardtPNS16, DBLP:conf/www/ZafarVGG17, DBLP:conf/ijcnn/DaiJHZW24, DBLP:conf/ijcnn/JiangDW23}, a field focused on developing techniques and methodologies to ensure ML systems make decisions that are both accurate and equitable across different population subgroups.
Fair ML encompasses various approaches \cite{DBLP:conf/nips/HardtPNS16, DBLP:conf/www/Wu0W19,DBLP:conf/icml/ZemelWSPD13, DBLP:conf/innovations/DworkHPRZ12,DBLP:conf/ijcai/Wu0W19, DBLP:conf/ijcai/ZhangWW17, DBLP:conf/nips/Wu0WT19,DBLP:journals/kais/KamiranC11,DBLP:conf/aistats/ZafarVGG17, DBLP:conf/www/ZafarVGG17,DBLP:conf/icml/XianY023, DBLP:conf/nips/PetersenMSY21}, from mathematical frameworks for quantifying fairness to practical techniques for mitigating bias throughout the machine learning pipeline.

Although recent research has made strides in fairness-aware machine learning, putting these research advances into practice remains difficult.
The field offers a wealth of theoretical frameworks, fairness metrics, and algorithmic solutions, but applying them in real-world settings often requires specialized expertise. Navigating the complexities of fairness, from choosing the right metrics to implementing bias mitigation techniques, can be daunting and present significant barriers to adoption.

To address these challenges, we present FairAgent, an automated system powered by large language models (LLMs) that streamlines the development of fairness-aware machine learning models.
FairAgent enables practitioners to interact with the entire ML pipeline, from data analysis and preprocessing to model training and evaluation, using an intuitive web interface.
By harnessing the capabilities of LLMs, FairAgent can automatically detect potential biases in datasets, suggest suitable fairness metrics and mitigation strategies, appropriately manage sensitive attributes, and optimize models for both accuracy and fairness.
The core contributions of FairAgent are:
(1) automated data preprocessing and bias detection via an LLM-driven data analysis module,
(2) automated model building and hyperparameter tuning that jointly optimize for accuracy and fairness, and
(3) a user-friendly interface that lowers the barrier to entry for fairness-aware machine learning, even for those without deep technical expertise.
Extensive experiments show that FairAgent achieves significant performance and meets fairness requirements across a range of fairness metrics, while reducing development effort and expertise requirements.
By automating complex technical decisions and providing guided, accessible workflows, FairAgent marks a significant advance toward democratizing fairness-aware machine learning.

\section{Related work}

\noindent\textbf{Fairness-aware ML.}
The field of fairness-aware machine learning has yielded a wide array of strategies to address algorithmic bias.
Significant research has established formal fairness metrics such as \textit{demographic parity}~\cite{DBLP:conf/sac/PedreschiRT12,DBLP:journals/csur/MehrabiMSLG21}, \textit{equalized odds}~\cite{DBLP:conf/nips/HardtPNS16}, causal fairness~\cite{DBLP:conf/uai/AgarwalLZ21}, and counterfactual fairness~\cite{DBLP:conf/nips/ZhangB18, DBLP:conf/aistats/MalinskyS019}, which have become standard benchmarks for evaluating fairness.
Building on these definitions, researchers have developed a variety of technical methods to achieve fairness in machine learning:
(1) \textit{pre-processing} techniques~\cite{kamiran2009classifying, DBLP:conf/icdm/CaldersKP09} that modify training data to mitigate inherent biases, for example, through reweighting or debiasing;
(2) \textit{in-processing} approaches~\cite{DBLP:conf/kdd/Corbett-DaviesP17,DBLP:conf/icdm/KamiranCP10,dblp:conf/pkdd/kamishimaaas12} that incorporate fairness constraints or regularization terms directly into the model training process;
(3) \textit{post-processing} methods~\cite{DBLP:conf/nips/HardtPNS16, dblp:conf/pkdd/kamishimaaas12} that adjust model predictions after training to better satisfy fairness criteria while maintaining predictive accuracy; and
(4) \textit{adversarial training} techniques that encourage the model to learn representations that obscure sensitive information.
Although these methods have demonstrated effectiveness in research environments, their practical implementation often demands significant expertise in both machine learning and fairness assessment, as practitioners must carefully balance trade-offs between model performance and competing fairness objectives.
This technical complexity and the need for meticulous tuning have hindered the widespread adoption of fairness-aware ML techniques in real-world applications.

\noindent\textbf{Automated machine learning.}
Automated machine learning (AutoML) has rapidly advanced as a powerful approach to simplifying the complexities inherent in traditional machine learning workflows.
AutoML platforms are designed to automate critical stages of the ML pipeline, such as feature engineering, model selection, and hyperparameter optimization, thereby reducing the need for specialized expertise.
Notable examples include AutoSklearn~\cite{feurer2015efficient}, which extends the scikit-learn library to deliver automated model selection and hyperparameter tuning using Bayesian optimization, as well as H2O AutoML~\cite{ledell2020h2o} and Auto-WEKA~\cite{DBLP:conf/kdd/ThorntonHHL13}, which strive to make machine learning more accessible to non-experts.
Despite their success in streamlining standard ML processes, these systems typically lack built-in mechanisms for fairness considerations, bias detection, or mitigation.
Moreover, most AutoML tools require input data to be preprocessed and properly formatted, which can present additional challenges for practitioners and further limit their accessibility.

\begin{figure}
    \centering
    \includegraphics[width=\linewidth]{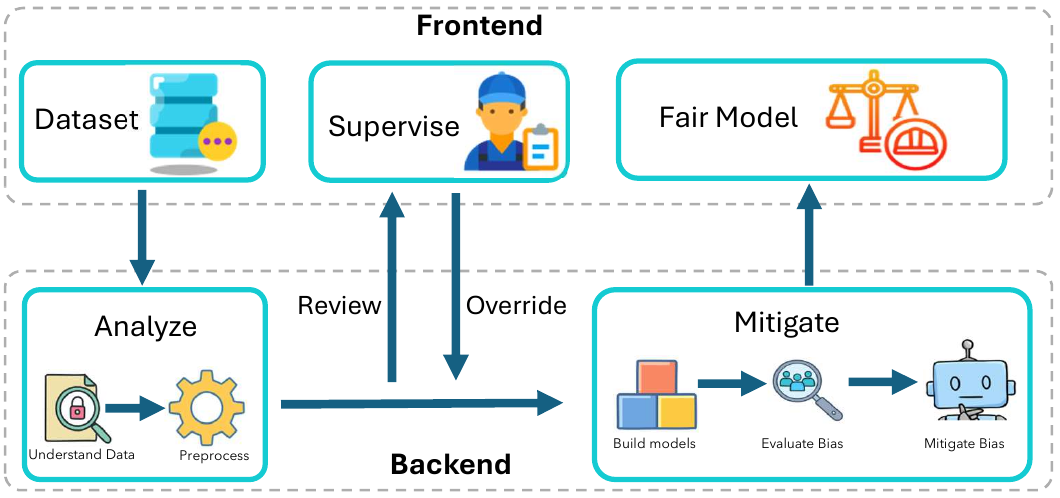}
    \caption{Overview of FairAgent.}
    \label{fig:systemdesign}
\end{figure}

\begin{figure*}[h]
    \centering
    \begin{subfigure}[b]{0.48\textwidth}
        \centering
        \includegraphics[width=\textwidth]{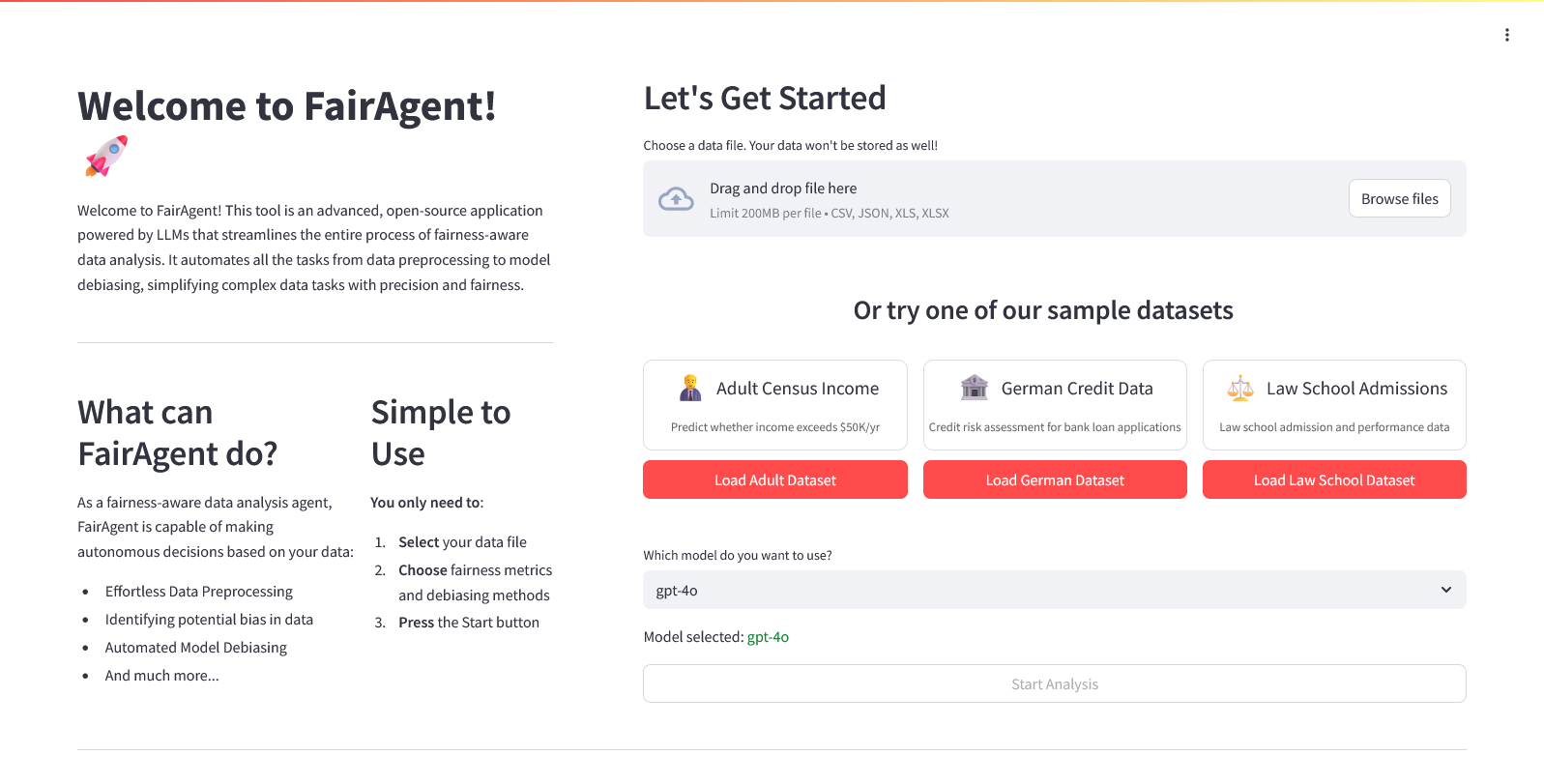}
        \caption{Dataset upload and LLM backbone selection}
        \label{fig:step1}
    \end{subfigure}
    \hfill
    \begin{subfigure}[b]{0.48\textwidth}
        \centering
        \includegraphics[width=\textwidth]{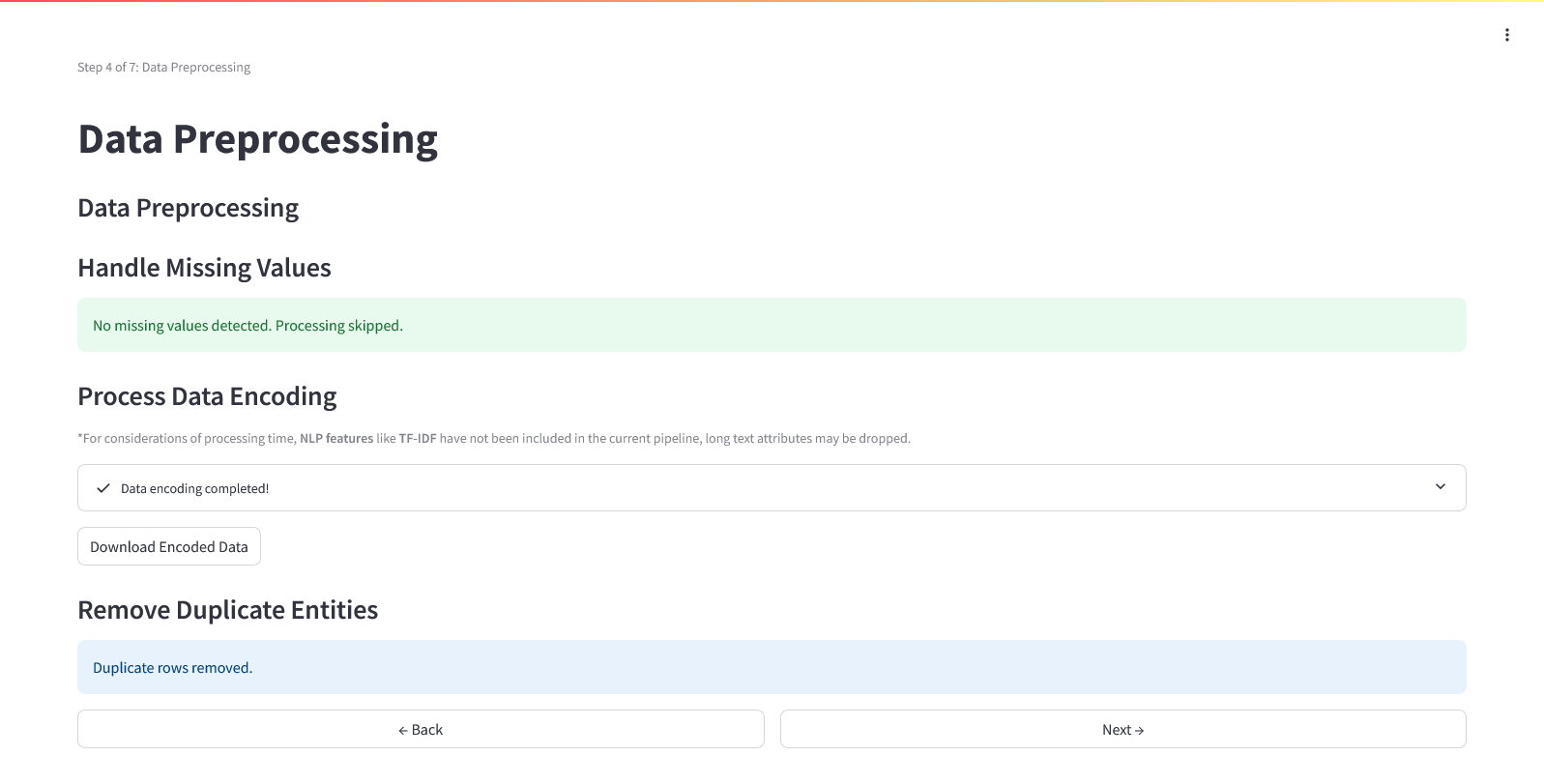}
        \caption{Data analysis and attribute configuration}
        \label{fig:step2}
    \end{subfigure}

    \vspace{0.5cm}

    \begin{subfigure}[b]{0.48\textwidth}
        \centering
        \includegraphics[width=\textwidth]{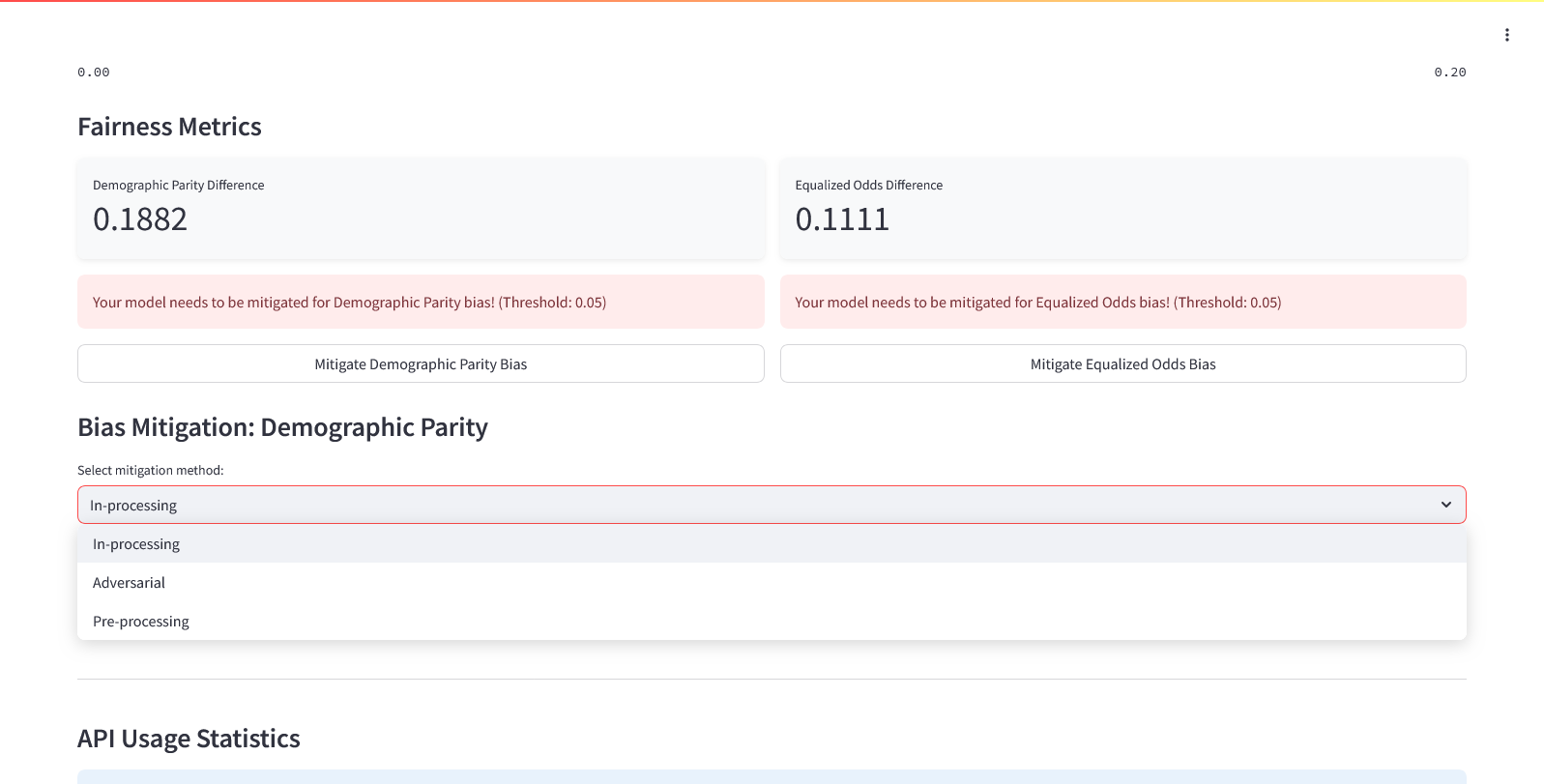}
        \caption{Bias evaluation and mitigation technique selection}
        \label{fig:step3}
    \end{subfigure}
    \hfill
    \begin{subfigure}[b]{0.48\textwidth}
        \centering
        \includegraphics[width=\textwidth]{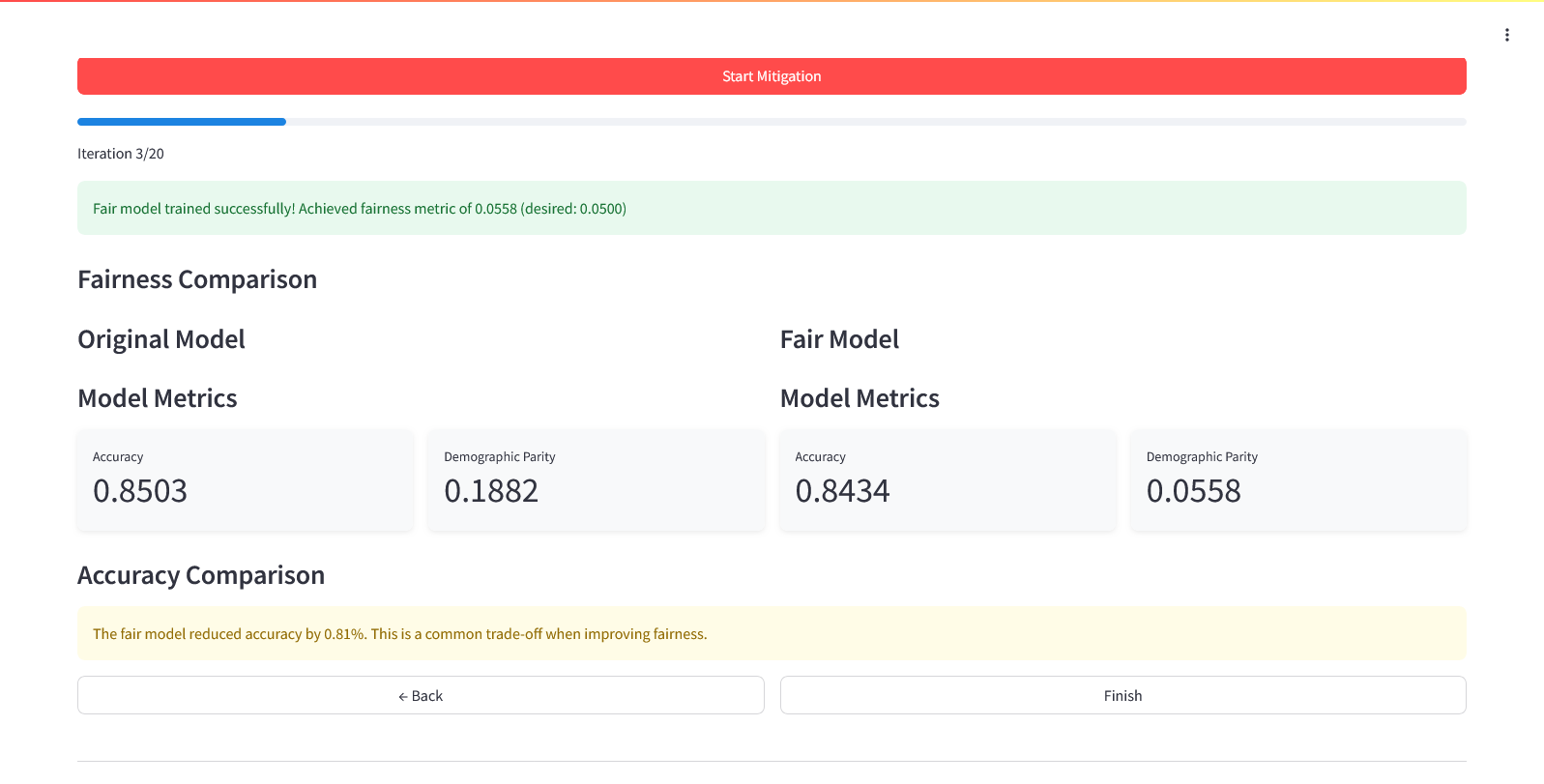}
        \caption{Result comparison between the vanilla and fair models}
        \label{fig:step4}
    \end{subfigure}
    \caption{System Walkthrough of FairAgent}
    \label{fig:walkthrough}
\end{figure*}

\section{System Design}
FairAgent\footnote{The demo is available at \url{http://tiny.cc/FairAgent}, and the source code and video recording are available at \url{http://tiny.cc/FairAgentCode}.} is implemented as a web-based application using the Streamlit framework. As illustrated in Figure~\ref{fig:systemdesign}, its architecture is composed of two primary components: (1) a user-friendly frontend interface that facilitates seamless interaction, and (2) a robust backend system that handles the core functionalities of FairAgent. These functionalities include data preprocessing, bias detection, and fairness-aware model training, all designed to ensure the system operates efficiently and effectively in reducing bias while maintaining optimal model performance.
In addition, a visual walkthrough of FairAgent is provided in Figure~\ref{fig:walkthrough}.

\subsection{Frontend Design}

The frontend interface is designed to streamline user input processes, allowing for seamless dataset uploads and task configurations.
Users have the flexibility to choose the large language model backbone that powers FairAgent, accept the suggested fairness metrics or select alternatives, and adopt or modify the proposed bias mitigation strategies.
Additionally, users can configure other optional parameters to suit their specific needs.
The system features an interactive supervision mechanism, enabling users to review and adjust automated decisions.
This capability is particularly beneficial when specific modifications are needed to override the automated procedure.
For example, users can manually specify sensitive and target attributes if the automated identification does not align with their analytical goals.

\subsection{Backend Design}

\noindent\textbf{Basic Analysis.}
Upon dataset upload, the system conducts a comprehensive automated analysis of the data by leveraging statistical methods to extract data insights.
This initial phase examines data dimensions, feature distributions, detailed feature descriptions with semantic interpretation, patterns of missing values across demographic subgroups, and correlation matrices highlighting potential issues.
This preliminary analysis serves dual purposes: (1) providing users with an immediate and comprehensive understanding of their dataset's characteristics, and (2) establishing a structured context with rich metadata that significantly enhances the LLM agent's ability to perform sophisticated fairness-aware processing.

\noindent \textbf{Contextual Analysis.}
The system performs a contextual analysis of the dataset, examining semantic relationships between features and their potential implications for fairness across different demographic groups.
This analysis incorporates knowledge about common fairness concerns, historical patterns of discrimination, and feature interactions that might amplify bias.
Based on this multifaceted analysis, the system generates informed recommendations for sensitive attributes (e.g., gender, race, age, disability status) and target attributes (predictions or outcomes).
Users can review these detailed recommendations, accept the suggestions as presented, or override them according to their domain knowledge and specific analytical requirements, with the system adapting its subsequent processing based on these user-defined parameters.

\noindent \textbf{Automatic Data Preprocessing.}
Following the contextual analysis, the system automatically preprocesses the data using insights gathered from both basic and contextual analyses.
The preprocessing pipeline applies appropriate transformations based on feature characteristics, such as implementing one-hot encoding for categorical variables and normalization for numerical features to prevent undue influence from extreme values.
For datasets containing missing values, the system employs sophisticated imputation strategies tailored to each feature's distribution and semantic meaning, including conditional mean imputation that accounts for demographic subgroups and pattern-based imputation that preserves existing relationships between features.

\noindent \textbf{Bias Detection and Baseline Model Training.}
To establish a reference point for fairness evaluation, the system trains conventional machine learning models such as logistic regression on the preprocessed dataset, selecting the best-performing model as the baseline.
These baseline models undergo rigorous evaluation across multiple dimensions, including predictive accuracy and comprehensive fairness assessments using demographic parity or equalized odds.
This comprehensive assessment quantifies existing biases in standard modeling approaches when applied to the given dataset, providing a crucial benchmark for subsequent fairness interventions.

\noindent \textbf{Fair Model Building, Fine-Grained Control and Tuning.}
FairAgent trains fairness-aware models that precisely meet user-specified fairness requirements while maximizing predictive performance, offering precise control over the fairness-accuracy trade-off.
The system first constructs a fairness-aware model using the mitigation method selected by the user (pre-processing techniques like reweighing or disparate impact remover, in-processing methods such as adversarial debiasing or prejudice remover, or post-processing approaches like reject option classification), derived from previous analyses and optimized for the specific dataset characteristics.
FairAgent then employs an advanced hyper-parameter tuning mechanism that systematically optimizes the model, exploring the hyper-parameter space efficiently to satisfy the user's fairness constraints as precisely as possible while maintaining maximum accuracy.
During this process, the system provides visualizations of the optimization trajectory, allowing users to observe how different parameter configurations affect both fairness and performance metrics.
Finally, the system presents a detailed comparative analysis between the fairness-aware model and the vanilla baseline, explicitly demonstrating improvements in both fairness metrics and the accuracy-fairness trade-off through interactive visualizations.

\begin{figure*}[!tbh]
    \centering
    \small
    \begin{minipage}{\textwidth}
    \begin{minipage}{0.48\textwidth}
    \centering
    \begin{table}[H]
    \centering
    \caption{Performance of FairAgent with Adversarial Training Configuration and Varying LLM Backends}
    \label{tab:ad_performance}
    \begin{tabular}{lcccccc}
    \toprule
    \multirow{2}{*}{\textbf{Backbone}} & \multirow{2}{*}{\textbf{Dataset}} & \multicolumn{2}{c}{\textbf{DP Mitigation}} & \multicolumn{2}{c}{\textbf{EO Mitigation}} \\
            &         & \textbf{Acc.} & \textbf{DP} & \textbf{Acc.} & \textbf{EO} \\ \midrule
    \multirow{2}{*}{\textit{Gemini-2.0}} & Adult            &     0.8385    &    0.0470   &     0.8415     &    0.0521     \\
    & Law         &      0.7250         &     0.0497       &    0.8954      &     0.0455    \\ \midrule
    \multirow{2}{*}{\textit{GPT-o1}} & Adult            & 0.8359       & 0.0456 &     0.8403     &  0.0546       \\
    & Law         &      0.9016         &    0.0598       &   0.8994       &    0.0406     \\ \midrule
    \multirow{2}{*}{\textit{GPT-4o}} & Adult            & 0.8176       & 0.0497 &    0.8478      &    0.0453     \\
    & Law         &   0.8986      &    0.0483        &    0.8984     &     0.0514    \\ \midrule
    \multirow{2}{*}{\textit{Claude-3.7}} & Adult            & 0.8224       & 0.0427 &     0.8454     &  0.0510       \\
    & Law         &   0.8986          &  0.0483         &     0.9056    &   0.0479   \\ \midrule
    \multirow{2}{*}{\textit{Deepseek-v2}} & Adult            & 0.8382     & 0.0450 &    0.8429      &   0.0576      \\
    & Law         &     0.9072        &  0.0536     &    0.9010     &    0.0533    \\ \midrule

    \end{tabular}
    \end{table}
    \end{minipage}%
    \hfill
    \begin{minipage}{0.48\textwidth}
    \centering
    \begin{table}[H]
    \centering
    \caption{Performance of FairAgent with Constrained Learning Configuration and Varying LLM Backends}
    \label{tab:in_performance}
    \begin{tabular}{lcccccc}
    \toprule
    \multirow{2}{*}{\textbf{Backbone}} & \multirow{2}{*}{\textbf{Dataset}} & \multicolumn{2}{c}{\textbf{DP Mitigation}} & \multicolumn{2}{c}{\textbf{EO Mitigation}} \\
            &         & \textbf{Acc.} & \textbf{DP} & \textbf{Acc.} & \textbf{EO} \\ \midrule
    \multirow{2}{*}{\textit{Gemini-2.0}} & Adult            & 0.8435        & 0.0535      &   0.8503       &    0.0550     \\
    & Law         &       0.9002       &    0.0512       &    0.9008    &  0.0508       \\ \midrule
    \multirow{2}{*}{\textit{GPT-o1}} & Adult            & 0.8440        & 0.0526 &      0.8506   &    0.0480    \\
    & Law         &      0.9013        &    0.0523        &    0.8968      &    0.0513     \\ \midrule
    \multirow{2}{*}{\textit{GPT-4o}} & Adult            & 0.8419       & 0.0479 &    0.8501      &    0.0548     \\
    & Law         &   0.9008           &    0.0518        &    0.9000      &    0.0548     \\ \midrule
    \multirow{2}{*}{\textit{Claude-3.7}} & Adult            & 0.8431       & 0.0482 &    0.8498      &    0.0469    \\
    & Law         &   0.9010            &  0.0515          &    0.8994      &    0.0527     \\ \midrule
    \multirow{2}{*}{\textit{Deepseek-v2}} & Adult            & 0.8419     & 0.0482 &     0.8495     &   0.0554      \\
    & Law         &      0.9000        &       0.0515     &    0.9005      &   0.0513      \\ \midrule

    \end{tabular}
    \end{table}
    \end{minipage}

    \begin{minipage}{\textwidth}
    \centering
    \begin{figure}[H]
    \centering
    \begin{subfigure}{0.24\textwidth}
        \centering
        \includegraphics[width=\linewidth, height=0.215\textheight]{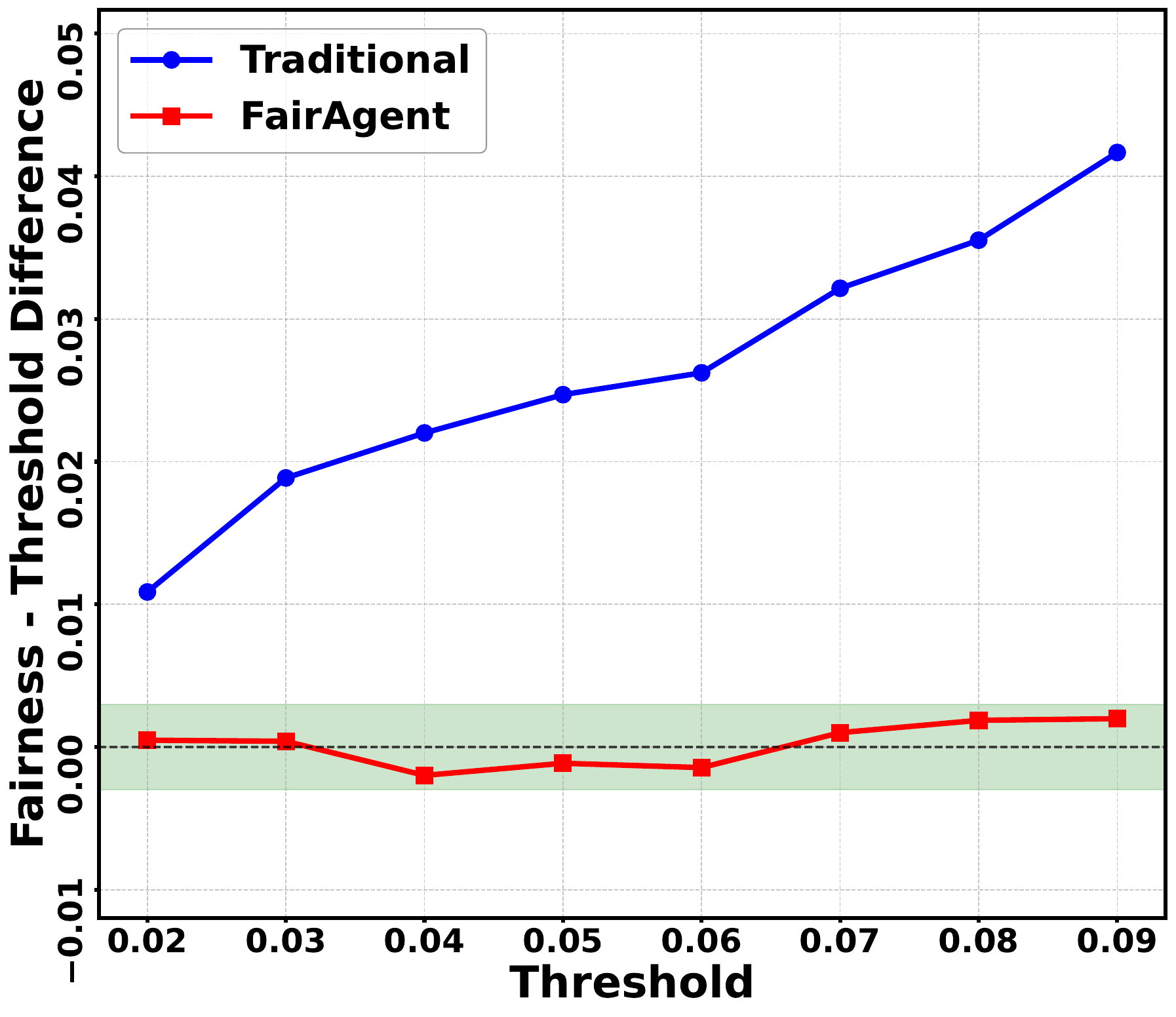}
        \caption{Enforcing DP on Adult Dataset}
        \label{fig:adult-rd}
    \end{subfigure}
    \hfill
    \begin{subfigure}{0.24\textwidth}
        \centering
        \includegraphics[width=\linewidth, height=0.215\textheight]{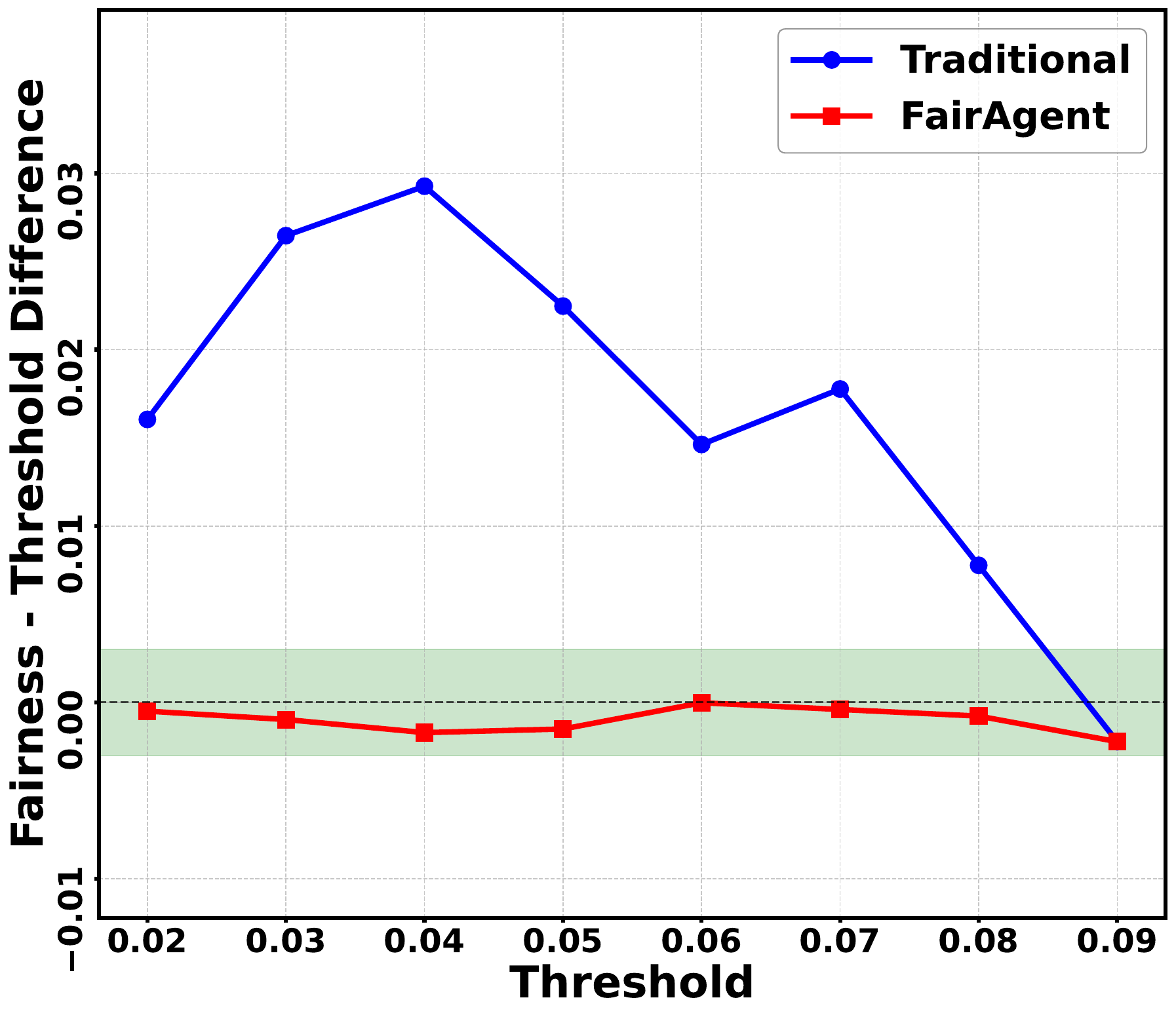}
        \caption{Enforcing EO on Adult Dataset}
        \label{fig:adult-eo}
    \end{subfigure}
    \hfill
    \begin{subfigure}{0.24\textwidth}
        \centering
        \includegraphics[width=\linewidth, height=0.215\textheight]{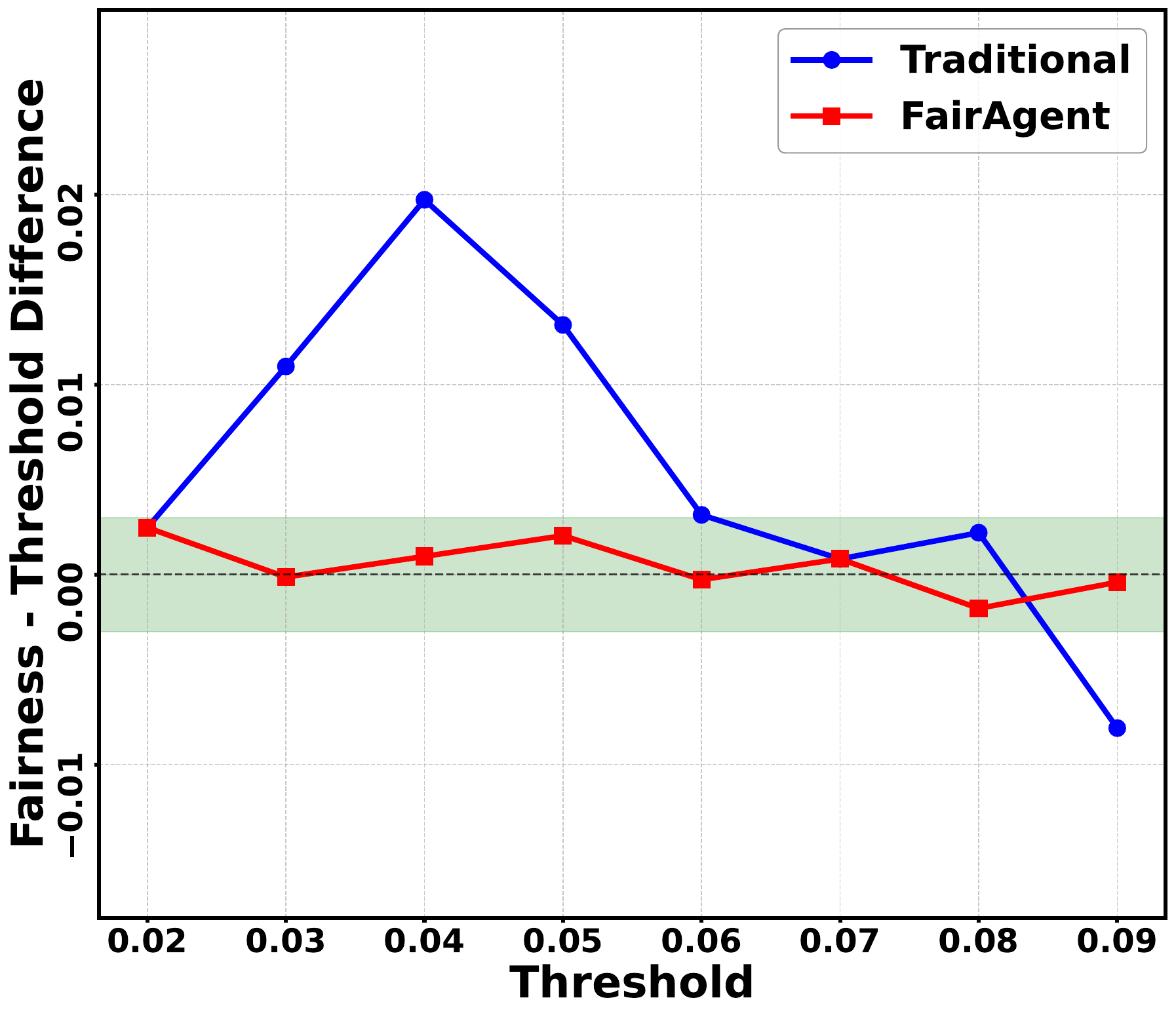}
        \caption{Enforcing DP on Law Dataset}
        \label{fig:law-rd}
    \end{subfigure}
    \hfill
    \begin{subfigure}{0.24\textwidth}
        \centering
        \includegraphics[width=\linewidth, height=0.215\textheight]{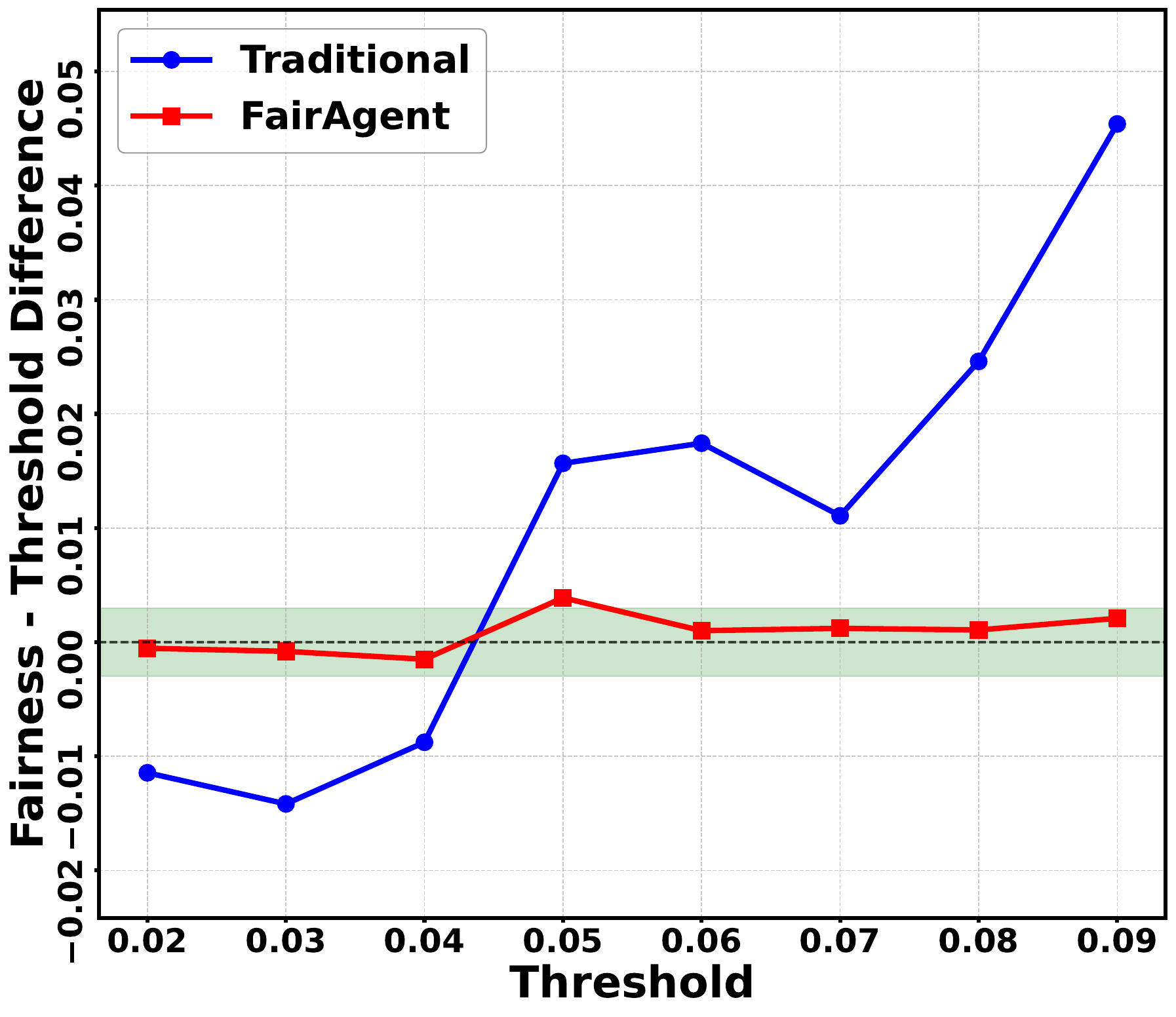}
        \caption{Enforcing EO on Law Dataset}
        \label{fig:law-eo}
    \end{subfigure}
    \caption{Fine-Grained Fairness Control at a Given Threshold for Demographic Parity and Equalized Odds}
    \label{fig:all}
    \end{figure}
    \end{minipage}
    \end{minipage}
    \end{figure*}

\section{Experiments}

\noindent \textbf{Datasets.}
We evaluate FairAgent on two widely recognized benchmark datasets in fairness-aware machine learning research.
The \textbf{Adult} \cite{adult_2} dataset comprises $48,842$ instances with 14 features, including demographic attributes (age, sex), socioeconomic indicators (education, occupation), and household characteristics.
The prediction task involves classifying whether an individual's annual income exceeds \$50,000.
The \textbf{Law School} \cite{wightman1998lsac} dataset, compiled by the Law School Admission Council (LSAC), tracks the academic progression of approximately 27,000 law students from 1991 to 1997, documenting their enrollment, graduation outcomes, and bar examination performance.
The prediction task is to predict whether a law student will pass the bar examination.
In the experiments, we set the fairness threshold as 0.05 with a tolerance of 0.01.

\subsection{Performance and Fairness Evaluation}
To evaluate FairAgent's effectiveness in mitigating bias, we conduct experiments on mitigating two widely-used fairness metrics: demographic parity (DP) and equalized odds (EO).
It is worth noting that other fairness metrics, such as statistical parity or disparate impact, can be easily integrated into FairAgent by simply changing the fairness metric in the code.
As shown in Table~\ref{tab:ad_performance} and Table~\ref{tab:in_performance}, our approach significantly reduces both demographic parity and equalized odds disparities across all datasets while preserving good accuracy.
These findings demonstrate that FairAgent effectively balances model performance with the given fairness threshold.

\subsection{Fine-Grained Fairness Control}
By integrating sophisticated fairness-aware control with advanced model tuning mechanisms, FairAgent ensures that the trained model's fairness metrics align precisely with user-defined fairness requirements, e.g., a threshold of 0.05.
In this section, we vary the fairness threshold from 0.02 to 0.09 and report the differences in fairness metrics between the trained model and the given fairness threshold.
As demonstrated in Figure~\ref{fig:all}, FairAgent outperforms conventional approaches with a smaller difference (around $\pm 0.005$) across two datasets and two metrics.
This capability provides fine-grained control over fairness thresholds, making FairAgent particularly valuable for applications requiring strict adherence to fairness standards.

\section{Conclusion}
In this paper, we introduced FairAgent, an LLM-powered automated system that democratizes fairness-aware machine learning by eliminating the need for deep technical expertise through automated bias detection, data preprocessing, and model optimization that jointly balances accuracy and fairness requirements.
Our approach addresses critical challenges in algorithmic fairness by providing a no-code solution that enables precise control over fairness objectives while maintaining model performance.
Through comprehensive evaluations across multiple datasets and backbone models, we demonstrated FairAgent's ability to efficiently achieve user-defined fairness constraints with significantly reduced development effort and expertise requirements.
As algorithmic fairness becomes increasingly critical, FairAgent represents a significant step toward democratizing responsible AI development through accessible, automated fairness-aware machine learning.

\section{Acknowledgments}

This work was supported in part by NSF 2520496, 2242812, and SC EPSCoR 24-GA02.

\bibliographystyle{IEEEtran}
\providecommand{\noopsort}[1]{}

\end{document}